# Expediting Building Footprint Extraction from High-resolution Remote Sensing Images via Progressive Lenient Supervision

**Haonan Guo[1], Bo Du[3,4,5], Chen Wu[1], Xin Su[*2], Liangpei Zhang[1]**

1 State Key Laboratory of Information Engineering in Surveying, Mapping and Remote Sensing, Wuhan University, Wuhan, China

2 School of Remote Sensing and Information Engineering, Wuhan University, Wuhan, China

3 National Engineering Research Center for Multimedia Software, Wuhan University, Wuhan, China

4 Institute of Artificial Intelligence, Wuhan University, Wuhan, China

5 School of Computer Science, Wuhan University, Wuhan, China

## Abstract

The efficacy of building footprint segmentation from remotely sensed images has been hindered by model transfer effectiveness. Many existing building segmentation methods were developed upon the encoder-decoder architecture, in which the encoder is finetuned from the newly developed backbone networks that are pre-trained on ImageNet. However, the heavy computational burden of the existing decoder designs hampers the successful transfer of these modern encoder networks to remote sensing tasks. Even the widely adopted deep supervision strategy fails to mitigate these challenges due to its invalid loss in hybrid regions where foreground and background pixels are intermixed. In this paper, we conduct a comprehensive evaluation of existing decoder network designs for building footprint segmentation and propose an efficient framework denoted as BFSeg to enhance learning efficiency and effectiveness. BFSeg achieves consistent and high performance gain with little computational cost across multiple backbones. Specifically, a densely connected coarse-to-fine feature fusion decoder network that facilitates easy and fast feature fusion across scales is proposed. Moreover, considering the invalidity of hybrid regions in the down-sampled ground truth during the deep supervision process, we present a lenient deep supervision and distillation strategy that enables the network to learn proper knowledge from deep supervision. Building upon these advancements, we have developed a new family of building segmentation networks, which consistently surpass prior works with outstanding performance and efficiency across a wide range of newly developed encoder networks.

## 1. Introduction

Building footprint extraction, which aims to assign each pixel in a remotely sensed image with a semantic category of building and non-building, plays a crucial role in remote sensing image processing(Dold and Groopman, 2017). Automatic building footprint segmentation holds significant value for various building-related geospatial applications, including sustainable urban planning, automatic driving, emergency response, and 3D urban modeling(Anderson et al., 2017; Su et al., 2021; Zhu et al., 2012). Consequently, precise automatic building footprint extraction has aroused considerable attention over the past decades(Masoumi and van Genderen, 2023).

Among the existing building extraction methods, deep learning-based algorithms have emerged as a prominent research area in recent years(Jiao et al., 2023). As compared with conventional methods that rely heavily on handcrafted features, the deep learning-based convolution neural network(CNN) can automatically learn building features from remote sensing images and make end-to-end predictions(X. Liu et al., 2022). The encoder-decoder architecture, characterized by its symmetrical and simple design, has found widespread adoption in deep learning-based building extraction(Ronneberger et al., 2015). The encoder part extracts hierarchical features from the input image, while the decoder part refines the spatial resolution of features and generates pixel-wise building predictions. Typically, the encoder part is fine-tuned from feature extractors pretrained on ImageNet, such as ResNet(Guo et al., 2021b; He et al., 2015; Xu et al., 2018) and VGGNet(Guo et al., 2021c; Ji et al., 2019; Simonyan and Zisserman, 2015; Wei et al., 2020; Zhang et al., 2020). With the recent advance of feature extraction networks in computer vision tasks such as Transformers (Dosovitskiy et al., 2021; Liu et al., 2021; Ze Liu et al., 2022) and convolutional networks(Zhuang Liu et al., 2022; Tan and Le, 2020), it is feasible to employ and transfer these well-developed feature extractors pretrained on large-scale natural image datasets to the remote sensing task for superior feature presentations. However, although these natural image-pretrained models can provide a fundamental understanding of images, there exists a large difference between natural images and remote sensing images.

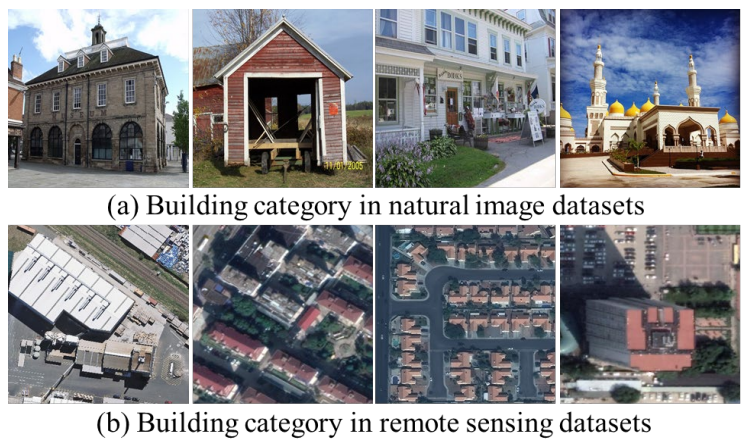

(a) Building category in natural image datasets

(b) Building category in remote sensing datasets

**Fig 1**. Illustration of the domain gap between building categories in natural image and remote sensing datasets.

As illustrated in Fig.1, the significant disparities in context and texture information between the two image domains, particularly for the building category, highlight the domain gap. Unfortunately, many existing building footprint segmentation decoder designs are ineffective in transferring the natural image-pretrained encoder networks to the building segmentation task due to their ineffective feature fusion designs and learning diagrams. In Fig.2, we examine the complexity and performance gain of the existing decoder designs and identify two primary challenges:

**Challenge 1:** Feature fusion- As the recently-developed feature extractors possess better feature representation ability and a larger number of output channels, decoders developed based on U-Net(Zhu et al., 2017), such as MA-FCN(Ji et al., 2019) and DS-Net(Zhang et al., 2020), bring an excessive computation burden in the decoder part. For example, as demonstrated in Fig.2, the usage of advanced encoder networks such as Swin Transformer(Liu et al., 2021) and ConvNeXt(Zhuang Liu et al., 2022) in the U-Net architecture results in an approximate 40% increase in computational cost within the decoder part. Such excessive computational burden hampers the effective transfer of the encoder network to specific remote sensing tasks due to the gradient disappearing problem. Some lightweight decoders, such as UPerNet(Xiao et al., 2018), bring about the loss of the model's accuracy despite its low computational cost. A lightweight and effective decoder is required to fully harness the outstanding learning potential of these newly designed backbone networks.

**Challenge 2**: Model learning- A common solution for addressing the gradient disappearing phenomenon that hampers transfer effectiveness is the deep supervision technique, which involves down-sampling the ground truth to specific resolutions and calculating the loss function from intermediate layer outputs. However, we have observed that the boundary regions of the deeply supervised predictions consistently exhibit low confidence and high loss values throughout the training process. This can be attributed to the inaccuracies in the boundary regions of the down-sampled ground truth. With such inaccurate ground truth boundaries, the network can neither learn from incorrect boundary pixels nor boundary information of the upper scales. Hence, it is vital to explore novel approaches to handle these impure down-sampled boundary pixels, which haven't been studied by the existing building footprint extraction methods.

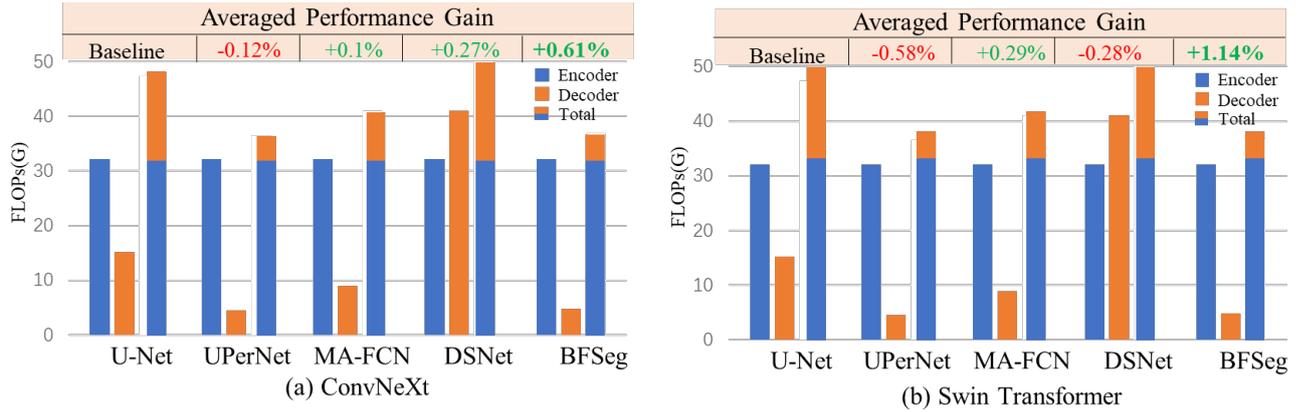

Fig 2. Comparison of complexity and performance gains of the existing decoder designs.

In this paper, we propose an efficient building footprint segmentation framework (BFSeg) to address the abovementioned problems. To mitigate the heavy computational burden associated with decoder design, we propose a lightweight and effective decoder network that facilitates the seamless transfer of newly developed encoder networks to remote sensing interpretation tasks. BFSeg achieves consistent and high performance gain with little computational cost across multiple backbones, as illustrated in Fig.2, indicating that it can be applied to any advanced backbones. In BFSeg, features from deep convolutional layers are densely connected to shallow layers, enabling shallow features to focus only on the interested area by progressively refining coarse predictions to the finer ones. Furthermore, we encourage the network to progressively refine deep prediction under the guidance of semantic information, in comparison to the existing methods that simply treat deep supervision as an auxiliary task. Last but not least, to alleviate the model difficulties caused by improper deep supervision, we propose a lenient deep supervision and self-distillation strategy that helps the model learn from proper regions. The main contributions are summarized as follows:

➢ We propose a densely connected coarse-to-fine feature fusion framework that facilitates the easy and fast transfer of well-designed backbone networks to the building footprint extraction task. Our framework surpasses existing frameworks in performance and computational cost, demonstrating exceptional effectiveness and efficiency.

➤ We propose a novel lenient deep supervision strategy to address the invalid model learning issues in boundary areas arising from conventional deep supervision. The strategy mitigates the problem by providing more lenient guidance during the deep supervision process in the training phase.

➤ Building upon the lenient deep supervision, we extend the approach to lenient self-distillation. It enables deep features to learn not only from ground truth labels but also from valuable knowledge from upper scales, and thus enhance the model's learning capacity.

➤ We demonstrate our method through experiments in several large-scale building footprint datasets under different encoder networks, showing the superiority of the proposed method over the existing architectures.

## 2. Related Works

### 2.1 Semantic Segmentation

Traditional semantic segmentation methods aim to extract manually designed features to distinguish foreground objects from the background. The extracted features are then classified into different semantic categories on a per-pixel basis(Bouziani et al., 2010; Huang et al., 2011). Ongoing development of traditional semantic segmentation involves designing improved feature descriptors and incorporating novel classifiers to achieve enhanced accuracy (Awrangjeb et al., 2013; Huang and Zhang, 2011). However, manually selecting the optimal feature descriptor is time-consuming and cannot meet the requirement of generalization limited by expert knowledge(Guo et al., 2021a).

In contrast to traditional handcrafted approaches, deep learning-based methods have the capacity to automatically learn optimal descriptors and perform end-to-end classification (Guo et al., 2022). Stacked convolution neural layers serve as automatic image descriptors that are able to learn hierarchical feature representation through backpropagation(Krizhevsky et al., 2017). To further enhance the performance of convolution neural networks(CNN), various endeavors have been made, including increasing model capacity(Simonyan and Zisserman, 2015), introducing residual learning(He et al., 2015), boosting multiscale perception(Szegedy et al., 2015, 2014), exploring neural architecture search(Tan and Le, 2020), and harnessing vision transformers(Dosovitskiy et al., 2021; Liu et al., 2021; Ze Liu et al., 2022). However, since these methods were initially developed for patch-wise classification, additional structures are necessary to refine feature resolution and enable pixel-wise classifications.

The fully convolutional network(FCN) structure was proposed to enable pixel-wise predictions by replacing fully connected layers with deconvolution layers(Shelhamer et al., 2017), which progressively fuses multiscale features and refines feature resolution. Further improvements have been made in increasing feature resolution and model perception fields (Chen et al., 2016) or aggregating global context information(Zhao et al., 2017). The encoder-decoder architecture, initially employed in U-Net(Ronneberger et al., 2015) and SegNet(Badrinarayanan et al., 2016), is a more sophisticated architecture. Compared to early methods that directly generate predictions using deep features, the encoder-decoder architecture progressively restores feature resolution by combining encoder features from various scales. This integration allows for the preservation of both high-level semantics and low-level details. Developing upon this architecture, RefineNet designed a multi-path refinement module to effectively fuse multiscale features and enhance high-resolution feature representation(Lin et al., 2016). Deeplab v3+ (Chen et al., 2017) proposed an atrous spatial pyramid pooling module to aggregate high-level semantic information in the encoder network and refine feature details using the decoder architecture, surpassing the former versions of the Deeplab series. Recently, with the remarkable progress of transformer networks in computer vision, some studies have replaced convolutional encoder networks with transformers, such as Swin Transformer (Chen et al., 2021; Liu et al., 2021), with UperNet(Xiao et al., 2018) as the decoder.

In these encoder-decoder model designs, a prevalent practice involves utilizing newly developed feature extraction networks as encoders and initializing their parameters with models pretrained on large-scale datasets like ImageNet. These networks are then fine-tuned for specific segmentation tasks by propagating gradients from the decoder network. Although this approach works well for natural downstream tasks, where the pre-trained model and the segmentation model are trained on natural images, it fails to bridge the domain gap between pre-trained models and remote sensing tasks. Consequently, a more refined architecture is required for geospatial segmentation tasks, allowing for effective and efficient transfer of pre-trained models trained on natural images to remote sensing tasks.

## 2.2 Building Footprint Extraction

Most deep learning-based remote sensing segmentation methods follow the basic architecture of the encoder-decoder. However, improvements are needed to tailor this architecture to the specific characteristics of remote sensing images. For instance, some studies have focused on addressing the intra-class variance present in remote sensing images by introducing foreground-background relationships and latent contrastive information(Yang and Ma, 2022; Zheng et al., 2020). In the context of building footprint extraction, Xu et al. enhanced the original U-Net architecture by incorporating a ResNet encoder to improve building feature representation(Xu et al., 2018). Wei et al. proposed an MA-FCN architecture that integrates multiscale building predictions generated at different stages of the U-Net

decoder(Wei et al., 2020). DSNet feeds the input image into two parallel branches for global-local building feature extraction; the extracted global information is utilized to guide and refine local feature representation(Zhang et al., 2020). Similarly, MAP-Net incorporates multiple parallel paths to preserve both global context information and local-detailed information, utilizing an attention module to enhance discriminative channels, and employing a spatial pyramid module to capture global dependencies(Zhu et al., 2020). Guo et al. introduce a U-Net-based coarse-to-fine boundary refinement network that effectively refines the boundary regions of the extracted buildings(Guo et al., 2022). Many existing building footprint segmentation networks adopt the U-Net architecture(Guo et al., 2021b, 2021c; Ji et al., 2019; Wei et al., 2020; Zhang et al., 2020), incorporating the deep supervision technique(Guo et al., 2022; Wei et al., 2020; Zhang et al., 2020). However, the conventional U-Net architecture with deep supervision suffers from insufficient feature fusion and ineffective model learning problems. On the one hand, the conventional U-Net decoder design introduces an excessive computation burden, making it difficult for the advanced backbone networks to exert their outstanding learning potential due to the gradient disappearing problem. On the other hand, the deep supervision strategy faces challenges in learning from inaccurate regions in down-sampled building labels and risks overfitting to the boundary information of higher scales. Hence, it is crucial to develop new architectures and learning diagrams to achieve efficient and effective building footprint segmentation, thereby harnessing the exceptional learning capabilities of advanced foundational models.

## 3. Methodology

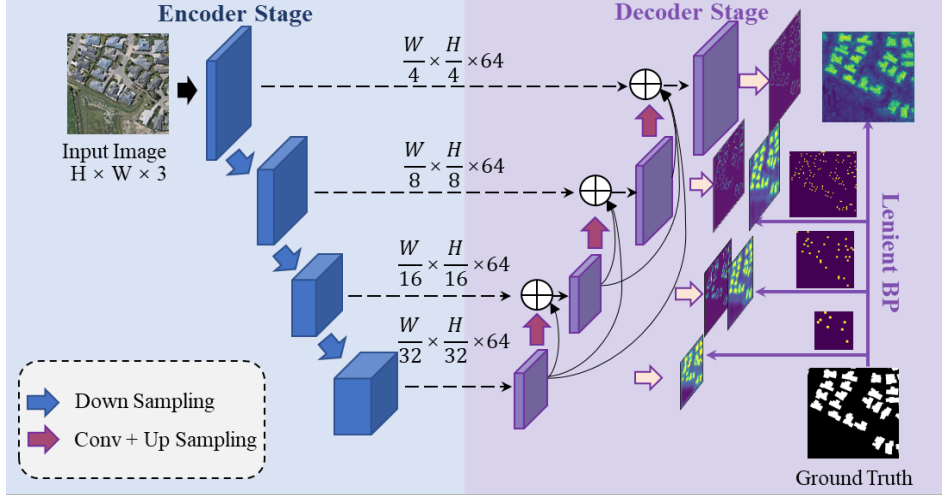

Fig 3. The overall architecture of the proposed BFSeg framework.

The proposed Efficient Building Footprint Segmentation network, BFSeg, is depicted in Fig.3. It comprises an encoder network of arbitrarily advanced feature extractors, a lightweight and effective feature pyramid network(LightFPN), and a lenient back-propagation(LenientBP) learning mechanism. BFSeg utilizes the proposed LightFPN module and achieves efficient refinement of building predictions in a coarse-to-fine manner. Moreover, a lenient learning strategy is designed to alleviate the issue of invalid object boundary learning during the deep supervision and distillation stage. We will introduce these modules in succession.

### 3.1 LightFPN

Given an input of a very high-resolution remote sensing image, the encoder network first extracts hierarchical image features of varying sizes; a decoder network is utilized to fuse the hierarchical features, resume feature resolution, and make pixel-wise building classification. Recent advancements in feature extraction networks, such as ConvNeXt(Zhuang Liu et al., 2022) and Swin Transformer(Liu et al., 2021), highlight the need to design a lightweight and effective decoder network to exert the representation potential of these advanced networks.

However, the widely-used decoder architecture of U-Net simply concatenates and convolutes hierarchical features, resulting in an excessive computational burden due to the large number of feature channels in the encoder features, as illustrated in Fig.4 (a). This ineffective decoder design incurs more than a third of additional computational costs and fails to optimize the feature extractors effectively. On the other hand, some lightweight decoders such as the UperNet, directly up-sample and aggregate deep features with shallow features transmitted from the encoder(Fig.4 (b)). Despite high computation efficiency, it may introduce noises from shallow features, as they encompass information about the entire image rather than specific regions of interest. To alleviate the issues of the computational burden and the noise issues associated with the existing decoder designs, we propose a novel decoder module denoted as LightFPN.

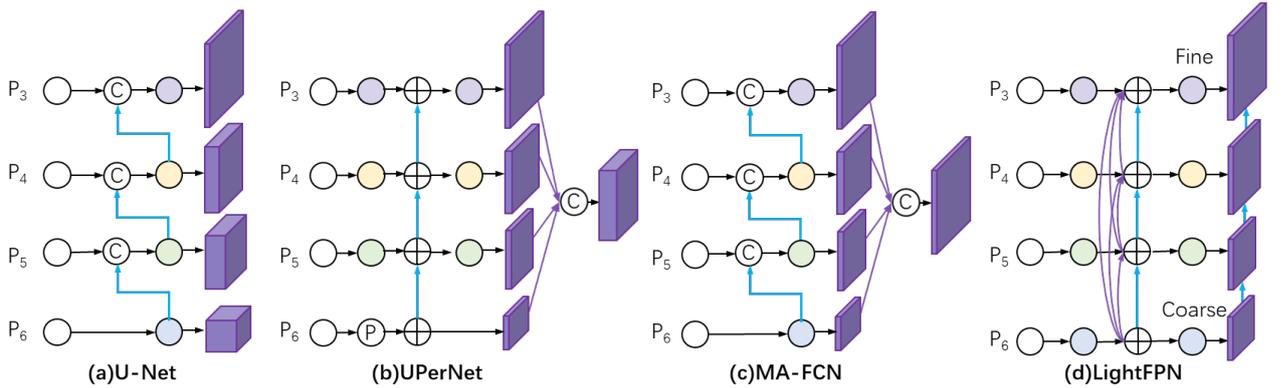

Fig 4. Comparison among different decoder designs for building footprint segmentation.

As depicted in Figure 4(d), our proposed LightFPN takes the hierarchical image features $P = \{P_3, P_4, P_5, P_6\}$ extracted by the encoder network as input. In feature set $P$, feature $P_i$ possess $\frac{1}{2^{i-1}}$ the spatial resolution of the input image and its feature channels are larger under lower spatial resolution. To account for the varying number of feature channels in $P$, a convolutional layer is first applied to each input feature to condense the number of feature channels to 64:

$$P' = \{Conv_{1x1}^{64}(P_3), Conv_{1x1}^{64}(P_4), Conv_{1x1}^{64}(P_5), Conv_{1x1}^{64}(P_6)\} \qquad (1)$$

where $Conv_{kxk}^{c}$ denotes the convolutional layer with kernel size $kxk$ and output channel $c$. Afterward, a classifier is applied to the lowest-resolution feature $P_6$ and generate prediction $Cls_6$. The subsequent stages are designed to capture and refine the residual component of the coarse prediction, integrating information from earlier decoder stages, forming a coarse-to-fine learning paradigm:

$$Cls_i = \delta(F_i) + Cls_{i-1} \qquad (2)$$

However, directly learning the residual prediction from shallow features, such as $P_i$ extracted by the encoder part(Islam et al., 2017), may introduce noise and undermine the quality of the final prediction due to limited semantic information in shallow layers. To this end, features from previous decoder stages are introduced to reconstruct shallow features and thus ensure the acquisition of high-resolution features with enhanced semantics:

$$F_i = Conv_{3x3}^{64}\big(\cup_{k=i-1}^{k} F_k \cup P_i'\big) + F_{i-1} \qquad (3)$$

Based on the reconstructed feature $F_i$, the residual classification term of scale i can be generated as:

$$\delta(F_i) = Conv_{3x3}^{1}(F_i) \qquad (4)$$

We then explain why LightFPN is a superior decoder architecture for building footprint segmentation. Firstly, the coarse-to-fine design of LightFPN enables the later decoder stage to focus on refining the spatial details of the coarse prediction. By leveraging deep semantic features, the irrelevant noise contained in shallow encoder features can be effectively suppressed, allowing the shallow classification results to accurately depict the residual term of the coarse prediction. Moreover, LightFPN boasts the advantage of being a lightweight decoder network that facilitates rapid and efficient feature fusion across various scales. This notably reduces the computational burden of the decoder component and enables seamless integration with the advanced encoder network for optimal performance in building footprint segmentation tasks.

## 3.2 Lenient supervision

Deep supervision is a critical technique in the encoder-decoder architecture for building footprint segmentation, alleviating the issue of gradient disappearing in the training process. In deep supervision, the high-resolution ground truth labels and the coarse model prediction are unified to the same spatial resolution using interpolation before calculating the deep supervision loss. However, during the interpolation process, some building boundary pixels become hybrid in the down-sampled ground truth due to the mixture of down-sampled pixels from building and nonbuilding pixels within the corresponding high-resolution area. However, the model faces difficulty in learning such hybrid boundary regions from the down-sampled ground truth. Specifically, when confronted with a bilinearly down-sampled pixel containing multiple categories within its corresponding high-resolution pixel, accurately predicting the proportion of each ground truth class becomes challenging, as the task degenerates into a sub-pixel problem. Conversely, the nearest neighbor-interpolated ground truth introduces uncertainty to these hybrid pixels during the learning process. To address this challenge, we propose a lenient deep supervision strategy to ignore the hybrid down-sampled pixels during the deep supervision learning process. In our detailed implementation, given building ground truth map $y \in [0,1]$ where 0 denotes the non-building pixels and 1 denotes the building pixels, we generate mask $m$ of the impure down-sampled pixels at each down-sampled scale. This is achieved by calculating the floor and ceil values of the bilinearly down-sampled ground truth map. Pixels with consistent values are pure regions for loss calculation; pixels with inconsistent values are impure regions that are ignored in the backpropagation process:

$$y_{down} = Inter(y) \tag{5}$$

$$m = floor(y_{down}) * ceil(y_{down}) \qquad (6)$$

where $Inter(\cdot)$ denotes bilinear interpolation that downsamples the building ground truth. The deep supervision loss in the masked areas is set to 0 and thus stops the model from learning from the hybrid regions that cause network uncertainty. In this way, the network focuses on learning from the pure pixels and thus facilitates model convergence by learning from the proper pixels. In implementations, the formula for applying lenient deep supervision is:

$$L_{linient} = \sum_{i=0}^{4} \frac{1}{\sum m}(CE(\hat{y}_i, y_{down}) \cdot m) \qquad (7)$$

where $CE(\cdot)$ denotes the cross entropy loss; $\hat{y}_i$ denotes the building prediction generated at scale $i$. We then explain why the proposed lenient deep supervision strategy is effective for building footprint segmentation. Conventional deep supervision technique directly calculates the cross-entropy loss between the predicted building map and the corresponding ground truth. However, the presence of hybrid pixels hampers the model learning process, leading to a decrease in the efficiency of model convergence. In the proposed lenient deep supervision strategy, regions with hybrid pixels are masked and do not influence the model learning process. As a result, the model learns exclusively from appropriate regions at the specific scale, enabling it to better capitalize on the benefits of the deep supervision strategy.

## 3.3 Lenient Self-Distillation

Self-distillation is an effective strategy for enhancing the model's resilience to label noise(Guo et al., 2022) and improving feature discriminative capabilities (Zhang et al., 2019) by distilling the knowledge acquired by deep convolutional layers into the shallower layers. We extend the proposed lenient deep supervision strategy to lenient self-distillation, thereby ensuring effective learning while avoiding any detrimental influence from hybrid regions during the distillation process. Specifically, pseudo labels are generated from the building predictions made by the final classifier. The knowledge of the final classifier is then distilled into the shallow layers by calculating the cross-entropy loss between the pseudo label and the prediction from deep supervision. Mask $m$ is utilized to filter out the impure regions that may deteriorate the distillation process. The formula is as follows:

$$L_{distill} = \sum_{i=0}^{4} \frac{1}{\sum m} (CE(\hat{y}_i, Inter(\sigma(\hat{y}))) \cdot m) \tag{8}$$

where $\sigma(\cdot)$ denotes the sigmoid function that generates the probability of building.

## 3.4 Model Details

The model was implemented using the Pytorch framework and was trained using an RTX3090 GPU. The initial learning rate was set to 0.001, which was subsequently reduced by 30% of its current value if the accuracy did not improve over three consecutive epochs. The model parameter optimization was carried out using the AdamW(Loshchilov and Hutter, 2019) optimizer with an L2-regularizer of 5e-4 weight decay. To prevent overfitting, we adopted online data augmentation including rotation, vertical flipping, and horizontal flipping. Each image patch had a 25% probability of being rotated or flipped randomly. During training, the loss function was composed of the building prediction loss between the final prediction $\hat{y}$ and the building ground truth $y$, the lenient deep supervision loss as described in Eq.7 and the lenient self-distillation loss as described in Eq.8, which can be calculated as follows:

$$L_{total} = CE(\hat{y}, y) + L_{lenient} + L_{distill} \tag{9}$$

The model's performance was evaluated on the test dataset after it converged on the training set.

## 4. Experiments and Analysis

## 4.1 Experiments Settings

### 4.1.1 Datasets

The performance of the model was assessed using three large-scale building datasets, including the WHU building dataset(Ji et al., 2019), the DeepGlobe dataset(Demir et al., 2018), and the Typical City Building dataset("A dataset of building instances of typical cities in China," n.d.). These datasets, publicly accessible, encompass a wide array of buildings from various regions and scenes, making them appropriate for evaluating the effectiveness of the models in real-world mapping scenarios. In the following sections, we will provide a detailed introduction to each of these datasets. We will introduce these datasets in succession.

1) WHU Building Dataset

The WHU building dataset is a large-scale building footprint segmentation dataset with more than 220,000 buildings in Christchurch, New Zealand. The dataset covers varied residential, countryside, and industrial areas. The spatial resolution of the aerial images is 0.3m. The building polygons are manually checked and edited to ensure their high quality. The images and labels are cropped into 512x512 pixels patches. The training, validation, and test datasets are from distinct regions, and contain 4736, 1036, and 2416 patches, respectively. Please refer to (Ji et al., 2019) for detailed information on the WHU building dataset.

2) DeepGlobe Dataset

The DeepGlobe dataset is a global building footprint extraction dataset that covers distinct areas in Las Vegas, Paris, Shanghai, and Khartoum. The satellite images are captured by the WorldView-3 sensor with a spatial resolution of 0.3m. Since buildings are varied in sizes, materials, and styles across different cities, it poses a great challenge to the model's robustness to buildings of different scenes. As suggested by (Zhu et al., 2020), we divide the dataset into 5860, 979, and 2944 patches as the training set, the validation set, and the test dataset with a ratio of 6:1:3. Please refer to (Demir et al., 2018) for more details of the DeepGlobe dataset.

3) Typical City Building Dataset

The Typical City building dataset is a large-scale building dataset that contains 7260 samples of more than 63,000 building instances in the cities of Beijing, Shanghai, Shenzhen, and Wuhan, China. It is suitable for testifying the models' robustness in national building mapping. The remote sensing images are collected from Google Earth. The dataset contains 5,985 image patches with their corresponding building labels for training and 1,275 patches for testing. We suggest readers refer to ("A dataset of building instances of typical cities in China," n.d.) for more details of the Typical City building dataset.

**4.1.2 Comparison Methods**

To evaluate the effectiveness of the proposed framework, we compare BFSeg with ten state-of-the-art semantic segmentation or building extraction models, including BOMSC-Net(Zhou et al., 2022), MSRF-Net(Zhao et al., 2023), LRAD-Net(Liu et al., 2023) , MA-FCN(Wei et al., 2020), DSNet(Zhang et al., 2020), MAP-Net(Zhu et al., 2020) , HRNet(Sun et al., 2019), UperNet(Xiao et al., 2018), Deeplab v3+(Chen et al., 2018), and Res-U-Net(Xu et al., 2018).

**4.1.3 Evaluation Metrics**

After the model converged on the training dataset, we predicted the results on the test dataset and calculated the true-positive(TP), true-negative(TN), false-positive(FP), and false-negative(FN) between the predicted building map and the ground truth. The model's performance was then evaluated using four evaluation metrics, including precision, recall F1-score, and IoU. These metrics can be calculated as follows.

$$\text{Precision} = \frac{TP}{TP+FP} \tag{10}$$

$$\text{Recall} = \frac{TP}{TP+FN} \tag{11}$$

$$F1 = \frac{2 \times Precision \times \text{Recall}}{Precision + Recall} \tag{12}$$

$$\text{IoU} = \frac{TP}{TP+FN+FP} \tag{13}$$

where precision denotes the proportion of the correctly predicted buildings that are predicted as buildings; recall is the proportion of the correctly predicted buildings; F1-score is the harmonic mean value between the precision and the recall value.; IoU denotes the mean over the union between the predicted building map and the building ground truth.

## 4.2 Evaluation of the WHU Building Dataset

We first present the comparison results on the WHU building dataset. The evaluation outcomes for BFSeg and the comparison methods on the WHU dataset are displayed in Table 1. The highest, second-highest, and third-highest scores are denoted in bold, underlined, and red, respectively. We can see that the proposed BFSeg framework achieves the highest and second highest IoU and F1-score when equipped with ConvNeXt and Swin transformers, respectively. BFSeg demonstrates superior performance compared to the UPerNet architecture, surpassing it by 0.5% and 1.02% on IoU under the ConvNeXt and Swin Transformer backbones, respectively. This highlights the advantage of BFSeg in transferring the ImageNet-pretrained feature extractors to the building extraction task. Moreover, BFSeg outperforms the state-of-the-art building extraction frameworks, such as MAP-Net and DS-Net, by a margin of at least 0.3% on IoU. The quantitative evaluation results demonstrate the effectiveness of the proposed BFSeg framework compared to other comparison methods. To further evaluate the qualitative results, we visualize some examples of the building prediction results on the WHU building dataset in Fig.5, in which green, blue, and red represent true-positive, false-negative, and false-positive, respectively. From Fig. 5 rows 1-2 we can see that BFSeg successfully identifies all buildings, even those that exhibit similar characteristics to the background, using both the Swin Transformer and ConvNeXt backbones. Moreover, HRNet, MA-FCN, and MAP-Net omitted the long-size buildings due to inadequate consideration of global context information. Although DS-Net successfully depicts the long-size building, it requires twice the forward propagation steps compared to BFSeg, resulting in higher computational costs.

Table 1 Quantitative evaluation of the WHU Building dataset

| Method | IoU(%) | Precision(%) | Recall(%) | F1-score(%) |
|---|---|---|---|---|
| Deeplabv3+ | 89.43 | 94.31 | 94.53 | 94.42 |
| Res-U-Net | 89.31 | 94.4 | 94.31 | 94.35 |
| HRNet | 89.86 | 94.37 | 94.95 | 94.66 |
| MA-FCN | 90.18 | 94.75 | 94.92 | 94.83 |
| Swin-UPer | 89.87 | 94.96 | 94.38 | 94.67 |
| ConvNeXt-Uper | 90.45 | **95.55** | 94.43 | 94.98 |
| DSNet | 89.82 | 94.79 | 94.49 | 94.64 |
| BOMSC-Net | 90.15 | 94.50 | 95.14 | 94.8 |
| MAP-Net | 90.59 | 95.32 | 94.81 | 95.07 |
| LRAD-Net | 88.89 | 94.21 | 91.12 | 92.47 |
| MSRF-Net | 90.16 | 94.97 | 94.68 | 94.82 |
| BFSeg(ConvNeXt) | **90.95** | 95.47 | 95.05 | **95.26** |
| BFSeg(SwinTran.) | 90.89 | 95.34 | **95.12** | 95.23 |

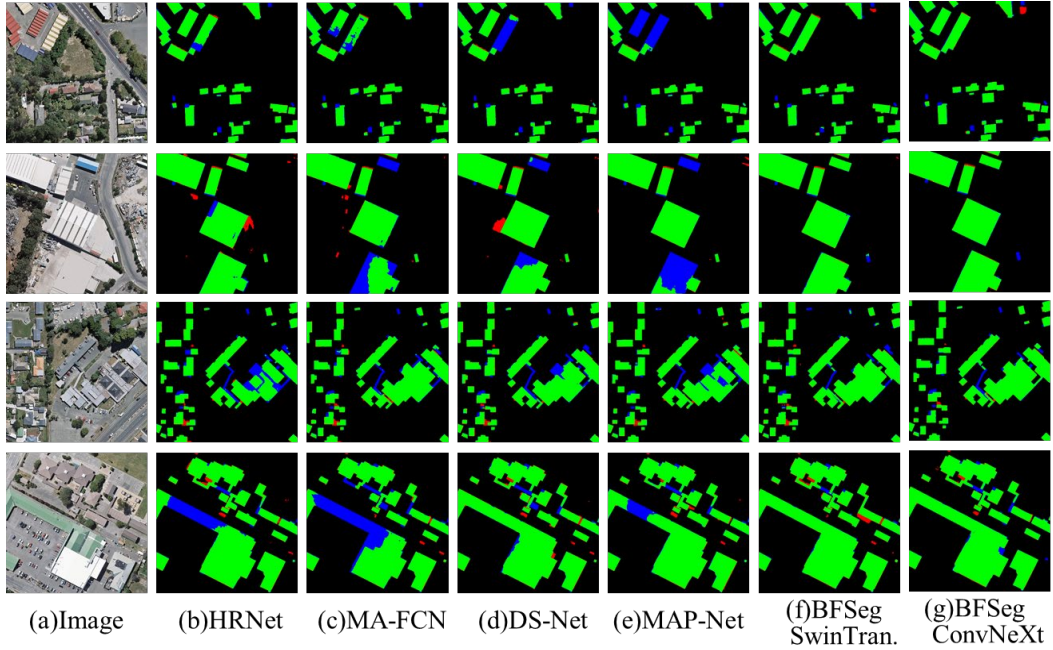

| (a)Image | (b)HRNet | (c)MA-FCN | (d)DS-Net | (e)MAP-Net | (f)BFSeg SwinTran. | (g)BFSeg ConvNeXt |

Fig 5.Visualization of building prediction results on the WHU Building dataset.

We conduct ablation experiments on the WHU building dataset to analyze the performance of the components in BFSeg. In the ablation experimental settings, we use the original U-Net architecture with ConvNeXt and Swin Transformer backbones as the baseline models. We then implement deep supervision and the proposed lenient supervision strategy alternately within the LightFPN framework to compare the efficacy of these components. It is evident from the results shown in Table 2 that the proposed LightFPN architecture outperforms the original U-Net architecture by 0.22% in terms of IoU under the ConvNeXt architecture and by 0.09% under the Swin Transformer architecture. With the original deep supervision strategy, the model's accuracy is improved by 0.3% and 0.9% under ConvNeXt and Swin Transformer backbones since it provides auxiliary supervisory in the learning process. It should be noted that the proposed lenient supervision strategy is a ready-to-use mechanism that further improves the models' performance by 0.37% and 0.92% under the ConvNeXt and Swin Transformer backbones with negligible computation cost. In conclusion, the experimental results affirm the effectiveness of the proposed components in BFSeg for effectively adapting prior CNN-based and Transformer-based feature extractors to the building extraction task.

Table 2 Ablation study on the WHU Building dataset

| Encoder | Decoder | Deep Supervision | Accuracy | | | |
|---|---|---|---|---|---|---|
| | | | IoU(%) | Precision(%) | Recall(%) | F1-score(%) |
| ConvNeXt | U-Net | × | 90.39 | 95.2 | 94.7 | 94.95 |
| | LightFPN | × | 90.61(+.22) | 95.37(+.17) | 94.78(+.08) | 95.07(+.12) |
| | LightFPN | +Ori. | 90.65(+.26) | 95.25(+.05) | 94.94(+.24) | 95.1(+.15) |
| | LightFPN | +Lenient | 90.86(+.47) | 95.38(+.18) | 95.05(+.35) | 95.21(+.26) |
| Swin Transformer | U-Net | × | 90.42 | 95.46 | 94.48 | 94.97 |
| | LightFPN | × | 90.51(+.09) | 95.07(-.39) | 94.96(+.48) | 95.02(+.05) |
| | LightFPN | +Ori. | 90.81(+.39) | 95.35(-.11) | 95.01(+.53) | 95.18(+.21) |
| | LightFPN | +Lenient | 90.83(+.41) | 95.35(-.11) | 95.05(+.57) | 95.2(+.23) |

## 4.3 Evaluation of the DeepGlobe Dataset

In addition, we conducted extensive experiments on the DeepGlobe dataset, with the quantitative evaluation and visualization results presented in Table 3 and Fig.6, respectively. BFSeg with ConvNeXt and Swin Transformer backbones achieves the highest and second-highest IoU and F1 score on the DeepGlobe dataset. Generally, the ConvNeXt backbone performs better than the Swin Transformer backbone on both BFSeg and UperNet architecture. BFSeg with the ConvNeXt backbone achieved an IoU of 81.46% and an F1-score of 89.78%, surpassing the state-of-the-art UPerNet and MAP-Net by 0.97% and 1.68% on IoU, respectively. The visualization results in Fig.6 also demonstrate the effectiveness of BFSeg. Fig.6 rows 1-2 reveal significant omissions and misclassifications in complex building structures for HRNet, MA-FCN, DS-Net, and MAP-Net, whereas BFSeg accurately captures building boundaries even under complex shapes. Furthermore, BFSeg excels at depicting small-scale buildings due to the advantage of multiscale feature fusion offered by LightFPN, as evidenced in Fig.6 row 4.

Table 3 Quantitative evaluation of the DeepGlobe Building dataset

| Method | IoU(%) | Precision(%) | Recall(%) | F1-score(%) |
|---|---|---|---|---|
| Deeplabv3+ | 76.12 | 87.1 | 85.79 | 86.44 |
| Res-U-Net | 78.46 | 88.11 | 87.75 | 87.93 |
| HRNet | 79.23 | 88.73 | 88.1 | 88.41 |
| MA-FCN | 79.08 | 88.78 | 87.86 | 88.32 |
| Swin-Uper | 78.05 | 88.57 | 86.79 | 87.67 |
| ConvNeXt-Uper | 80.49 | 89.87 | 88.52 | 89.19 |
| DSNet | 79.38 | 89.02 | 87.99 | 88.5 |
| MAP-Net | 79.78 | 89.66 | 87.86 | 88.75 |
| BFSeg(ConvNeXt) | **81.46** | **90.16** | **89.41** | **89.78** |
| BFSeg(SwinTran.) | 80.73 | 89.59 | 89.1 | 89.34 |

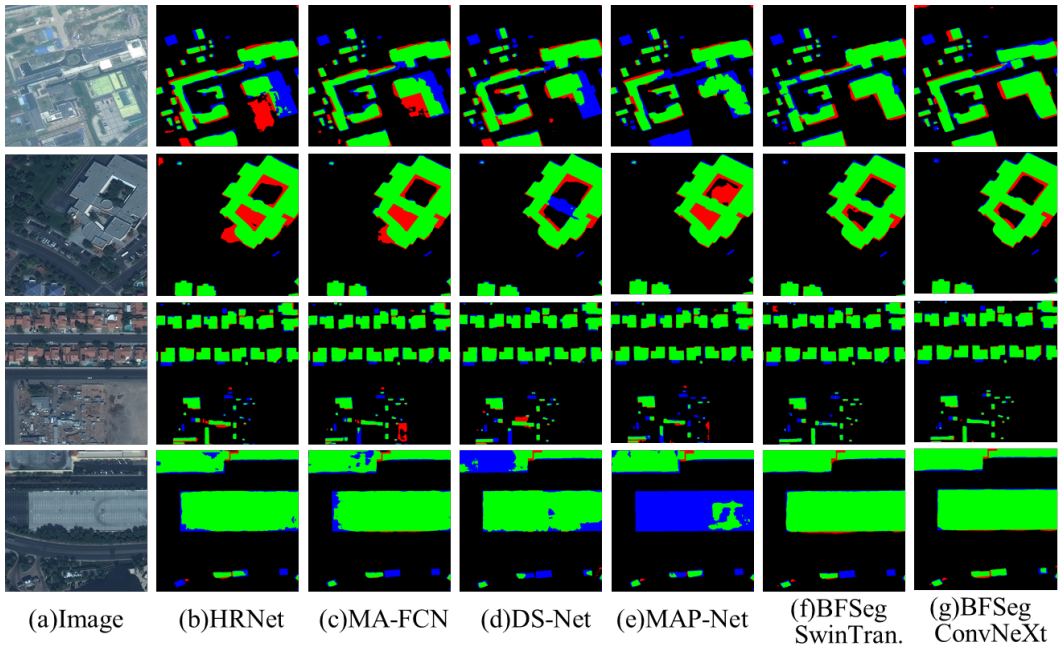

(a)Image  (b)HRNet  (c)MA-FCN  (d)DS-Net  (e)MAP-Net  (f)BFSeg SwinTran.  (g)BFSeg ConvNeXt

Fig 6.Visualization of building prediction results on the DeepGlobe dataset.

The ablation experimental results on the DeepGlobe dataset, as illustrated in Table 4, indicate that LightFPN yields greater improvement in model accuracy compared to U-Net under both ConvNeXt and Swin Transformer backbones. Moreover, the application of lenient supervision enhances the performance of the conventional deep supervision strategy on both backbones. With the combination of LightFPN and lenient supervision, BFSeg enables a seamless and efficient transfer of advanced feature extractors to the building footprint segmentation task.

Table 4 Ablation study on the DeepGlobe Building dataset

| Encoder | Decoder | Deep Supervision | Accuracy | | | |
|---------|---------|------------------|----------|--|--|--|
| | | | IoU(%) | Precision(%) | Recall(%) | F1-score(%) |
| ConvNeXt | U-Net | × | 80.87 | 89.75 | 89.11 | 89.43 |
| | LightFPN | × | 80.94(+.07) | 90(+.25) | 88.93(-.18) | 89.47(-.04) |
| | LightFPN | +Ori. | 81.37(+.5) | 90.02(+.27) | 89.44(+.33) | 89.73(+.3) |
| | LightFPN | +Lenient | 81.48(+.61) | 90.01(+.26) | 89.58(+.47) | 89.8(+.37) |
| Swin Transformer | U-Net | × | 79.17 | 88.95 | 87.81 | 88.38 |
| | LightFPN | × | 80.09(+.92) | 89.38(+.43) | 88.51(+.7) | 88.94(+.56) |
| | LightFPN | +Ori. | 80.63(+1.46) | 89.81(+.86) | 88.75(+.94) | 89.28(+.9) |
| | LightFPN | +Lenient | 80.67(+1.5) | 89.91(+.96) | 88.7(+.89) | 89.3(+.92) |

## 4.3 Evaluation of the Typical City Building Dataset

The Typical City Building dataset is a large-scale building dataset that covers diverse urban and suburban scenes in China. It is particularly suitable for simulating real-world scenarios in national building footprint mapping. The quantitative and qualitative experimental results are shown in Table 5 and Fig.7, respectively. In Fig.7 row 1, there appear holes in the prediction of the large-scale building, indicating the comparison methods cannot effectively fuse hierarchical features and thus lead to insufficient feature representation. Moreover, Fig.7 rows 2-4 illustrate that the comparison methods struggle to accurately discriminate complex buildings of complex shapes from the background, which is probably because the comparison methods cannot well exert the potential of the advanced feature extractor. The numerical evaluation results in Table 5 also demonstrate the effectiveness of BFSeg, which exceeds the building extraction DS-Net by 5.16% and 3.36% on the IoU and F1-score, respectively. The improvements lie in the ability of easy and fast transfer of BFSeg, which can fully exert the potential of the advanced CNN-based and Transformer-based backbones. On the contrary, many comparison architectures cannot extract discriminative features from the input image, leading to misclassification and omission in the building extraction result.

Table 5 Quantitative evaluation of the Typical City dataset

| Method | IoU(%) | Precision(%) | Recall(%) | F1-score(%) |
|---|---|---|---|---|
| Deeplabv3+ | 74.93 | 84.84 | 86.52 | 85.67 |
| Res-U-Net | 76.18 | 87.31 | 85.66 | 86.48 |
| HRNet | 77.14 | 88.48 | 85.75 | 87.09 |
| MA-FCN | 75.74 | 87.85 | 84.59 | 86.19 |
| Swin-Uper | 75.82 | 87.01 | 85.49 | 86.25 |
| ConvNeXt-Uper | 77.15 | 87.69 | 86.52 | 87.1 |
| DSNet | 72.73 | 85.55 | 82.91 | 84.21 |
| MAP-Net | 76.06 | 86.77 | 86.04 | 86.41 |
| MSRF-Net | 72.86 | 84.67 | 83.94 | 84.30 |
| BFSeg(ConvNeXt) | **77.89** | **87.71** | **87.43** | **87.57** |
| BFSeg(SwinTran.) | 77.28 | 87.16 | 87.2 | 87.18 |

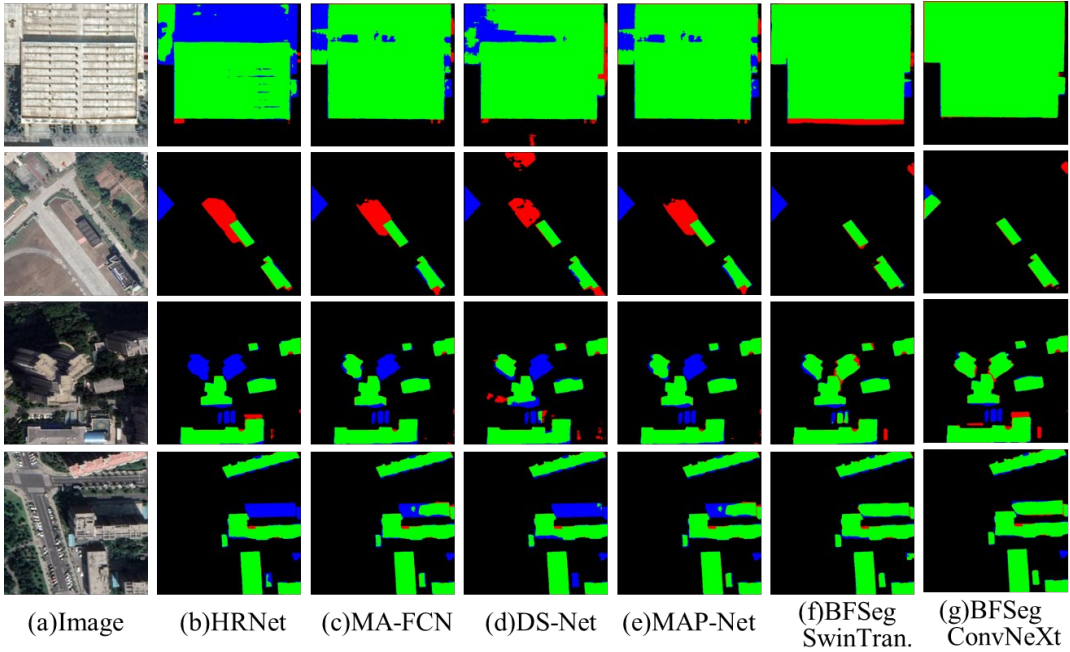

(a)Image　(b)HRNet　(c)MA-FCN　(d)DS-Net　(e)MAP-Net　(f)BFSeg SwinTran.　(g)BFSeg ConvNeXt

Fig 7.Visualization of building prediction results on the Typical City dataset.

The ablation experimental results in Table 6 indicate that LightFPN enhances the U-Net architecture with improvements of 0.34% in terms of IoU (with ConvNeXt backbone) and 0.99% IoU (with Swin Transformer backbone) on the China Typical Building dataset. It is worth noting that the application of the deep supervision strategy to the LightFPN architecture slightly decreases accuracy. This suggests that the hybrid regions in the down-sampled ground truth may impede feature learning, particularly when dealing with complex building structures. However, the proposed lenient supervision strategy can improve the model accuracy by masking the hybrid regions in the back-propagation process. The model can effectively learn from pure regions, enabling the extraction of discriminative representations of buildings.

Table 6 Quantitative evaluation of the Typical City building dataset

| Encoder | Decoder | Deep Supervision | Accuracy | | | |
|---------|---------|------------------|----------|----------|----------|----------|
| | | | IoU(%) | Precision(%) | Recall(%) | F1-score(%) |
| ConvNeXt | U-Net | × | 77.21 | 87.43 | 86.86 | 87.14 |
| | LightFPN | × | $77.55_{(+.34)}$ | $87.34_{(-.09)}$ | $87.37_{(+.51)}$ | $87.35_{(+.11)}$ |
| | LightFPN | +Ori. | $77.44_{(+.23)}$ | $87.52_{(+.09)}$ | $87.05_{(+.19)}$ | $87.29_{(+.15)}$ |
| | LightFPN | +Lenient | $77.62_{(+.41)}$ | $87.7_{(+.27)}$ | $87.1_{(+.24)}$ | $87.4_{(+.26)}$ |
| Swin Transformer | U-Net | × | 75.89 | 86.74 | 85.85 | 86.29 |
| | LightFPN | × | $76.88_{(+.99)}$ | $87.58_{(+.84)}$ | $86.29_{(+.44)}$ | $86.93_{(+.64)}$ |
| | LightFPN | +Ori. | $76.64_{(+.75)}$ | $87.1_{(+.36)}$ | $86.45_{(+.6)}$ | $86.77_{(+.48)}$ |
| | LightFPN | +Lenient | $76.92_{(+1.03)}$ | $87.61_{(+.87)}$ | $86.31_{(+.46)}$ | $86.95_{(+.66)}$ |

## 5. Discussion

In this section, we discuss the effectiveness of the lenient distillation strategy. We also compare and discuss the complexity and accuracy of different decoder designs.

### 5.1 The Effectiveness of Self-distillation

We evaluate the effectiveness of lenient self-distillation in Fig.8, from which we can see that in most cases, the proposed strategy can improve the models' feature representation ability by introducing knowledge of deep layers to guide the learning of shallow layers, except for the accuracy on the DeepGlobe dataset is slightly decreased when equipped with the ConvNeXt backbone. This observation suggests that the self-distillation strategy is a cost-effective method for improving the model's building extraction performance by distilling knowledge from deeper layers to shallower ones. The success of the lenient supervision and distillation strategy underscores the effectiveness of masking impure regions at higher scales during the processes of deep supervision and distillation. It enables the model to focus on learning from pure regions, leading to improved performance in building extraction tasks.

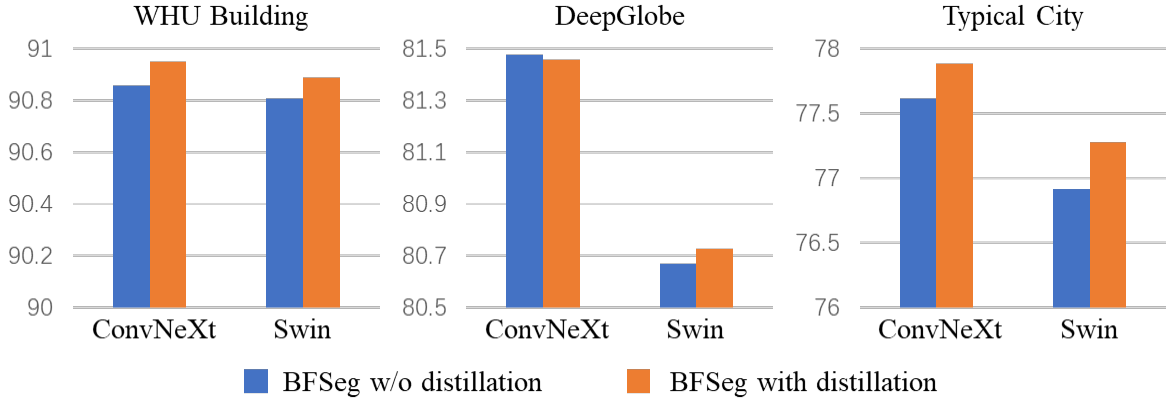

Fig 8. Comparison of BFSeg without/with lenient distillation strategy under three datasets.

## 5.2 Comparison of Decoder Designs

We compare the complexity and accuracy of different decoder designs using the advanced ConvNeXt and Swin Transformer backbones. The decoder architectures selected for comparison include U-Net, UPerNet, MA-FCN, and Dual-Stream architectures. We measure the complexity of these decoder designs using the number of parameters(Params) and floating point operations(FLOPs). The results presented in Table 7 indicate that the BFSeg framework, coupled with ConvNeXt and Swin Transformer, achieves the highest accuracy while maintaining the first or second least complexity. In comparison, although UPerNet is relatively lightweight, its accuracy is limited due to its inefficiency in transferring ImageNet-pretrained backbones to remote sensing tasks. In contrast, the Dual-Stream architecture of DS-Net showcases compatible performance in building extraction. However, its complexity is approximately tenfold higher than that of the proposed BFSeg framework, as it requires separate forward propagations of the input image for global-local feature extraction. DS-Net also significantly outperforms the original U-Net and MA-FCN architecture with an average of 8 times lower in the number parameter, 2 times lower in FLOPs, and higher accuracy with both ConvNeXt and Swin Transformer backbones. In general, BFSeg is a lightweight and effective framework for building footprint extraction that can be seamlessly integrated with various advanced feature extractors.

Table 7 Complexity and accuracy of different decoder choices for building footprint extraction

| Encoder | Decoder | Decoder Complexity | | WHU Building | | DeepGlobe | | Typical City | |
|---------|---------|--------|--------|--------|--------|--------|--------|--------|--------|
| | | Params(M) | FLOPs(G) | IoU(%) | F1(%) | IoU(%) | F1 (%) | IoU(%) | F1 (%) |
| ConvNeXt | U-Net | +6.53 | +16.35 | 90.39 | 94.95 | 80.87 | 89.43 | 77.21 | 87.14 |
| | UPerNet | +1.97 | **+4.45** | 90.45 | 94.98 | 80.49 | 89.19 | 77.15 | 87.1 |
| | MA-FCN | +6.31 | +8.91 | 90.49 | 95.01 | 81.21 | 89.63 | 77.08 | 87.06 |
| | Dual-Stream | +8.89 | +41.05 | 90.71 | 95.13 | 81.17 | 89.61 | 77.4 | 85.76 |
| | BFSeg | **+0.68** | +4.8 | **90.95** | **95.26** | **81.46** | **89.78** | **77.89** | **87.57** |
| Swin Transformer | U-Net | +11.51 | +25.56 | 90.42 | 94.97 | 79.17 | 88.38 | 75.89 | 86.29 |
| | UPerNet | +2.5 | **+4.85** | 89.87 | 94.67 | 78.05 | 87.67 | 75.82 | 86.25 |
| | MA-FCN | +18.35 | +8.52 | 90.7 | 95.13 | 79.91 | 88.83 | 75.74 | 86.19 |
| | Dual-Stream | +11.42 | +54.35 | 90.25 | 94.87 | 78.28 | 87.82 | 76.09 | 86.42 |
| | BFSeg | **+0.71** | +4.87 | **90.89** | **95.23** | **80.73** | **89.34** | **77.28** | **87.18** |

# 6. Conclusion

This article proposes BFSeg, a highly efficient framework designed for extracting building footprints from very high-resolution remote sensing images. BFSeg addresses the computational burden problem of U-Net design and the invalid learning problem of deep supervision. BFSeg effectively harnesses the exceptional learning capabilities of advanced natural image-pretrained backbones such as ConvNeXt and Swin Transformer. BFSeg incorporates a densely connected coarse-to-fine feature fusion decoder, enabling seamless and rapid feature integration across various scales. Additionally, a lenient learning strategy is implemented to mitigate the problem of invalid object boundary learning during the deep supervision and distillation stages. Built upon an enhanced network architecture and learning diagram, we have developed a new family of building segmentation networks utilizing advanced CNN-based and transformer-based backbones, such as ConvNeXt and Swin Transformer. Through extensive experiments on the WHU building dataset, the DeepGlobe dataset, and the Typical City dataset, BFSeg consistently outperforms previous network designs, boasting superior accuracy and efficiency. Moreover, its effectiveness extends beyond specific backbone as it proves to be applicable to a wide range of recently developed encoder networks, successfully transferring natural image-pretrained encoder networks to the building segmentation task. In future work, we will focus on designing foundation models for remote sensing building extraction by considering the characteristics of remote sensing images, aiming to enhance the effectiveness and accuracy of building extraction in practical scenarios.

**Acknowledgement**

This work was supported by National Natural Science Foundation of China under grant 62371348 and grant 42230108.